\documentclass{article}

\usepackage{arxiv}

\usepackage[utf8]{inputenc} 
\usepackage[T1]{fontenc}    
\usepackage{hyperref}       
\usepackage{url}            
\usepackage{booktabs}       
\usepackage{amsfonts}       
\usepackage{nicefrac}       
\usepackage{microtype}      
\usepackage{lipsum}
\usepackage{graphicx}
\graphicspath{ {./images/} }

\title{Topic Modelling: Going Beyond Token Outputs}

\author{
 Lowri Williams\\
  School of Computer Science \& and Informatics\\
  Cardiff University \\
  UK \\
  \texttt{WilliamsL10@cardiff.ac.uk} \\
   \And
 Eirini Anthi\\
  School of Computer Science \& and Informatics\\
  Cardiff University \\
  UK \\
  \texttt{AnthiES@cardiff.ac.uk} \\
  \And
 Laura Arman \\
  School of Social Science \\
  Cardiff University \\
  UK \\
  \texttt{ArmanL@cardiff.ac.uk} \\
  \And
 Pete Burnap \\
  School of Social Science \\
  Cardiff University \\
  UK \\
  \texttt{BurnapP@cardiff.ac.uk} \\
}

\begin{document}
\maketitle
\begin{abstract}
Topic modelling is a text mining technique for identifying salient themes from a number of documents. The output is commonly a set of topics consisting of isolated tokens that often co-occur in such documents. Manual effort is often associated with interpreting a topic's description from such tokens. However, from a human's perspective, such outputs may not adequately provide enough information to infer the meaning of the topics; thus, their interpretability is often inaccurately understood. Although several studies have attempted to automatically extend topic descriptions as a means of enhancing the interpretation of topic models, they rely on external language sources that may become unavailable, must be kept up-to-date to generate relevant results, and present privacy issues when training on or processing data. This paper presents a novel approach towards extending the output of traditional topic modelling methods beyond a list of isolated tokens. This approach removes the dependence on external sources by using the textual data itself by extracting high-scoring keywords and mapping them to the topic model's token outputs. To measure the interpretability of the proposed outputs against those of the traditional topic modelling approach, independent annotators manually scored each output based on their quality and usefulness, as well as the efficiency of the annotation task. The proposed approach demonstrated higher quality and usefulness, as well as higher efficiency in the annotation task, in comparison to the outputs of a traditional topic modelling method, demonstrating an increase in their interpretability.
\end{abstract}


\section{Introduction}
The increase of textual user-generated content on online platforms and social networks has become a key source for a wealth of information. Such platforms are often used to share information such as news, brands, political discussions, and more \cite{bakshy2012role}. Nevertheless, the large volume of text data that is broadly available on the web increases the challenge of identifying the most relevant information in real-time. Text mining is a promising solution to the problem of information overload associated with summarising and understanding vast amounts of unstructured text originating from diverse sources. More specifically, topic modelling aims to identify and extract salient concepts or themes, also known as ``topics'', distributed across a collection of documents \cite{kang2019analysis}.

Many studies have focused on extracting topics expressed in several domains, such as online social networks (e.g. \cite{curiskis2020evaluation}, \cite{chinnov2015overview}), particularly to identify influential individuals on a social media platform (e.g. \cite{weng2010twitterrank}) and detecting signs of depression in related language on Twitter (e.g. \cite{resnik2015beyond}), clinical applications such as triaging patients based on referral letters (e.g. \cite{spasic2020patient}), extracting scientific topics in academic journals (e.g. \cite{griffiths2004finding}), and most recently, understanding the public's opinion of governments during the COVID-19 pandemic (e.g. \cite{wright2021public}). 

The output from applying topic modelling to a collection of documents is commonly presented as a set of the top most co-occurring terms appearing in each topic \cite{jacobi2016quantitative}, \cite{greene2014many}. Manual effort is often associated with interpreting a topic's description from a set of isolated tokens, where a topic is often given a title or a name to reflect the understanding of its underlying meaning \cite{morstatter2018search}. For example, ``\textit{car, power, light, drive, engine, turn}'' may infer topics surrounding \textit{Vehicles}, and ``\textit{game, team, play, win, run, score}'' may infer \textit{Sports}. However, one of the key concerns with topic models lies with the subjectivity surrounding this task, as well as how well human readers can understand the topics, otherwise referred to as topic interpretability \cite{morstatter2018search}. More often than not, topic modelling outputs may not adequately describe the meaning of the topic itself. Problems include generic topic descriptions with too many words \cite{boyd2014care}, topics with disparate or poorly connected words \cite{mimno2011optimizing}, misaligned topics \cite{chuang2013topic}, and multiple nearly identical topics \cite{boyd2014care}. Subsequently, this misinterpretation leads to ineffective information retrieval and decision making.  

In this case, several studies have attempted to automatically extend topic descriptions, or select the best label for a given topic, as a means of enhancing the interpretation of topic models and reducing the human in the loop. Such approaches often rely on external textual sources (e.g. Wikipedia, WordNet) or sophisticated large language models (e.g. ChatGPT). In recent advancements, the development and maintenance of machine learning models and resources in privacy-aware settings have seen significant improvements, making the utilisation of external resources more feasible and secure than ever. However, despite these advancements, relying solely on such resources can still pose inherent risks and limitations. Firstly, the availability and stability of these resources remain a concern. Although less frequent, outages or restrictions in access to these external resources can disrupt the continuity and reliability of research relying on them. Moreover, the relevance and currency of information from these sources are not always guaranteed. In rapidly evolving fields, external resources might not be updated promptly, leading to the generation of outputs based on outdated information. Secondly, while privacy-aware machine learning models have made strides, they are not entirely foolproof. There are ongoing concerns regarding data breaches and the potential for these models to inadvertently memorise and reveal sensitive data. This is particularly crucial in contexts dealing with confidential or proprietary information, where the inadvertent leakage of data through these models could have serious ramifications.

Given these considerations, this paper proposes an approach which aims to strike a balance between leveraging the advancements in privacy-aware machine learning and mitigating the risks associated with dependency on external resources. By developing a methodology that primarily relies on internal data processing and analysis, this method aims to maintain a high degree of control over data relevance and privacy, ensuring that the model remains robust and reliable. More specifically, this paper proposes:

\begin{itemize}
    \item A novel approach towards extending the output of traditional topic modelling methods beyond token outputs. The initial experiments presented in this paper are tested using Latent Dirichlet Allocation (LDA), a traditional approach for distributing text segments into topics. However, given the positive findings of the experiments, and to demonstrate the generalisation of the proposed method, the approach is also applied when two of the latest state-of-the-art topic modelling methods, BERTopic and Top2Vec, are used.
    
    \item An evaluation of how such extended outputs improve the interpretability of topic descriptions from a human perspective.
\end{itemize} 

\begin{figure}
\centering
\includegraphics[width=0.5\textwidth]{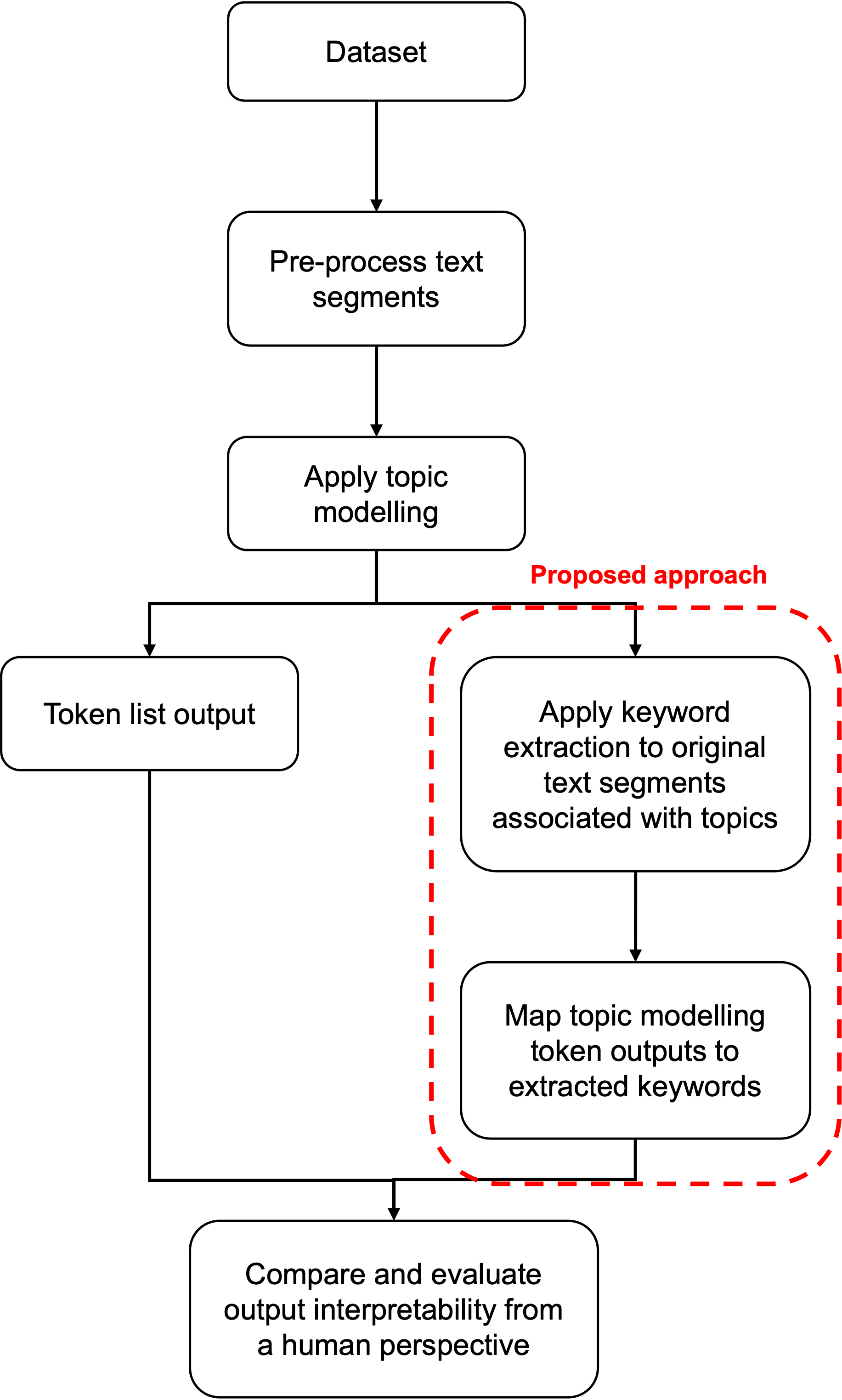}
\caption{An overview of the study design}
\label{study_design}
\end{figure}

The study was designed as shown in Figure \ref{study_design}: 1) pre-process text responses using traditional Natural Language Processing (NLP) techniques, 2) apply the topic modelling algorithm, 3) for each topic in 2, extend the token list output by applying keyword extraction to the original text segments, 4) map token outputs from 2 to the extended outputs in 3, and 5) evaluate the ease of interpreting outputs from 2 and 4 by human readers.

The remainder of this paper is structured as follows: Section \ref{sec:background} presents the related work, Section \ref{sec:datasets} discusses the selection of text corpora used to support the experiments herein and the techniques used to prepare the data for such experiments, Section \ref{sec:topic_modelling} discusses topic modelling and how it was applied to the datasets, Section \ref{sec:keyword_extraction} discusses the proposed methodology of using keyword extraction to extended topic modelling outputs, Section \ref{sec:evaluation} evaluates the ease of interpreting such outputs by human readers, Section \ref{sec:generalisation} discusses the generalisation of the proposed method by using other topic modelling methods, Section \ref{sec:generalisation_data} discusses the generalisation of the proposed method when using new data, and finally, Section \ref{sec:conclusion} concludes the paper.

\section{Related Work}\label{sec:background}

Several current works that have adopted topic modelling often rely on the manual approach of interpreting and determining the topic's main descriptive label based on the token results generated by the topic modelling approach (e.g. \cite{yu2021tourist}, \cite{kumari2021topic}). However, previous work has shown that using top terms are not enough for interpreting the coherent meaning of a topic following topic modelling \cite{mei2007automatic} \cite{hindle2013automated}. Such studies include Lee et al. \cite{lee2017human}, who qualitatively evaluated how readers understand, assess, and refine topics. They demonstrate a disconnect between how readers perceive topic model outputs and the operations that current interactive systems support. 
In this case, several studies have attempted to automatically extend topic descriptions as a means of enhancing the interpretation of topic models. Recent works have explored the use of external sources, such as Wikipedia, WordNet, or other ontologies for supporting the automatic labelling of topics by deriving candidate labels using lexical-based (e.g. \cite{lau2011automatic}, \cite{magatti2009automatic}, \cite{mei2007automatic}, \cite{bhatia2016automatic}, \cite{basave2014automatic}) or graph-based algorithms applied on these sources (e.g. \cite{aletras2014labelling}, \cite{hulpus2013unsupervised}). For example, recently, both Allahyari et al. \cite{allahyari2017knowledge} and Kinariwala and Deshmukh \cite{kinariwala2021onto_tml} propose an ontological approach for generating topic labels. Both approaches use the semantic relatedness of LDA outputs and their ontological categories to determine topic labels. However, limitations of this approach include the reliance of an ontology of terms, which in the case of \cite{kinariwala2021onto_tml}, is only 500 words which are categorised and limited to only four main domains (crime, environment, politics, and sports). Others, such as Bhatia et al. \cite{bhatia2016automatic}, have used word embeddings to map topic modelling outputs to Wikipedia article titles using cosine similarity and relative ranking measures. Whilst others, such as Bechara et al. \cite{bechara2021transfer}, propose a transfer learning approach for topic labelling which uses domain-specific codebooks to automatically label topics.

Although the aforementioned works propose methods of automatically assigning more informative and more generic labels based on topic models outputs, they lack emphasis on whether outputs from such approaches are interpretable and meaningful to human readers. To address this, other studies have evaluated their outputs by asking human annotators to label or score them based on how semantically related they were to a topic, how meaningful they are, and their usefulness and coherence. Wan and Wang \cite{wan2016automatic} propose an algorithm to extract text summaries that are much longer than keywords or phrases for describing topics. Aletras et al. \cite{aletras2017evaluating} compare three different topic representations in a document retrieval task and report that readers find phrased labels easier to interpret in comparison to a list of terms. Lau et al. \cite{lau2011automatic} also use phrases as topic labels, and propose a supervised approach for ranking candidate labels. In their work, candidate labels include the top five topic terms and noun chunks extracted from Wikipedia articles. Kou et al. \cite{kou2015automatic} propose another method of semantically mapping topics to candidate labels using word vectors and letter trigram vectors. Mei et al. \cite{mei2007automatic} propose to use n-grams for topic labelling and approach the problem of labelling as an optimisation problem which involves minimising the Kullback-Leibler divergence between word distributions and maximising the mutual information between a label and a topic model. Chang and Boyd-Graber \cite{chang2009reading} and Morstatter and Liu \cite{morstatter2018search} use crowdsourcing to measure topic interpretability using a numerical score. They validate topic models using word intrusion, an approach which requires participants to study the top words within a topic and to identify which words that do not belong. In addition, they also validate topics models using topic intrusion, an approach similar task to word intrusion, where participants study the topic probabilities for a document and evaluate how relevant they are to their understanding of the text. 

The work cited above focuses on extending topic modelling outputs, and more importantly, evaluating how interpretable such outputs are to human readers. However, such approaches use external textual sources that may be at risk of becoming unavailable. The size of such resources may also be relatively large, and therefore, generating results from them in real-time may be ineffective and computationally intense. There has yet to be an investigation into a lightweight approach that uses the textual data itself to extend topic modelling outputs for systems that require the almost immediate presentation of interpretable topics to humans.

\section{Text Corpora and Data Preparation}\label{sec:datasets}

To conduct the experiments described herein, the data was provided by our industrial research collaborator, a software company that focuses on the concept of crowdsourcing solutions to strategic business challenges. The eight datasets consist of short informal responses to a range of different questions, propositions, and requests for ideas. For example, dataset ID = 1 asked its participants, \textit{``Who does the channel reach, what is the purpose of the channel, and what information does the channel share?''}, whereas dataset ID = 4 asked, \textit{``What is your supply chain going to be, who are your most important customers, and how will you generate income?''}. Table \ref{tab:distribution} reports the distribution of responses across each dataset, with Table \ref{tab:preprocessing_text} reporting some examples of the text responses. By utilising these datasets, each representing unique variations and characteristics, the generality of the proposed approach of expanding topic modelling outputs can be tested, enhancing its reliability and real-world applicability. 

\begin{table}[ht]
\caption{Distribution of responses across datasets}
\label{tab:distribution}
\centering
\begin{tabular}{|c|c|c|c|}
\hline
\textbf{ID} & \textbf{Total Responses} & \multicolumn{1}{l|}{\textbf{Min \# of Tokens}} & \multicolumn{1}{l|}{\textbf{Max \# of Tokens}} \\ \hline
1                   & 84                       & 1                                                          & 48                                                         \\ \hline
2                   & 277                      & 1                                                          & 165                                                        \\ \hline
3                   & 156                      & 1                                                          & 102                                                        \\ \hline
4                   & 1,028                    & 1                                                          & 1,154                                                      \\ \hline
5                   & 149                      & 1                                                          & 105                                                        \\ \hline
6                   & 100                      & 1                                                          & 104                                                        \\ \hline
7                   & 149                      & 1                                                          & 109                                                        \\ \hline
8                   & 85                       & 1                                                          & 66                                                         \\ \hline
\end{tabular}
\end{table}

The data preparation and analysis in this study was conducted using Python (version 3.7.2). For text pre-processing, the traditional NLP techniques were applied, including: 

\begin{itemize}
    \item Converting text to lowercase.
    
    \item Removing additional white spacing using Python's regular expression package, RegEx (version 2020.9.27).
    
    \item Removing punctuation, digits, and emojis using RegEx. 
    
    \item Tokenising text using Python's natural language package, Natural Language Toolkit (NLTK) (version 3.4.1).
    
    \item Removing stop words as part of the NLTK package.
    
    \item Stemming tokens using the Porter Stemmer as part of the NLTK package.
\end{itemize}

Table \ref{tab:preprocessing_text} shows examples of the original text responses against how they are represented following pre-processing.

\begin{table}[ht]
\caption{Pre-processing text responses for topic modelling}
\label{tab:preprocessing_text}
\centering
\begin{tabular}{|c|l|l|}
\hline
\textbf{ID} & \multicolumn{1}{c|}{\textbf{Original Text Responses}}                                                                                                              & \multicolumn{1}{c|}{\textbf{Pre-processed Text Responses}}                                                                        \\ \hline
1                            & all staff                                                                                                                                                 & `staff'                                                                                                                  \\ \hline
2                            & \begin{tabular}[c]{@{}l@{}}improve staff moral and knowledge \\ of ongoing issues specifically within \\ specialist crime.\end{tabular}                   & \begin{tabular}[c]{@{}l@{}}`improv', `staff', `moral', `knowledg', \\ `ongo', `issu', `specialist', `crime'\end{tabular} \\ \hline
3                            & \begin{tabular}[c]{@{}l@{}}the army doesn't really have a lot \\ of equipment for the cold weather\end{tabular}                                           & `armi', `lot', `equip', `cold', `weather'                                                                                \\ \hline
4                            & \begin{tabular}[c]{@{}l@{}}most work would be submitted \\ electronically - I would need a \\ computer, screen, internet connection, \\ etc.\end{tabular} & \begin{tabular}[c]{@{}l@{}}`work', `submit', `electron', `comput', \\ `screen',  `internet', `connect'\end{tabular}      \\ \hline
5                            & \begin{tabular}[c]{@{}l@{}}mental health is a very important \\ aspect of staff wellbeing, and it \\ affects a lot of people.\end{tabular}                & \begin{tabular}[c]{@{}l@{}}`mental', `health', `aspect', `staff', \\ `wellb', `lot', `peopl'\end{tabular}                \\ \hline
6                            & \begin{tabular}[c]{@{}l@{}}working remotely, particularly in \\ transit, we need quick access to \\ colleagues contacts\end{tabular}                      & \begin{tabular}[c]{@{}l@{}}`working', `remot', `transit', `quick', \\ `access', `colleagu',  `contact'\end{tabular}      \\ \hline
7                            & \begin{tabular}[c]{@{}l@{}}would require some planning and\\ availability to be able to facilitate\\ such events.\end{tabular}                            & \begin{tabular}[c]{@{}l@{}}`requir', `plan', `avail', `facilit', \\ `event'\end{tabular}                                 \\ \hline
8                            & \begin{tabular}[c]{@{}l@{}}really beneficial to distance from \\ your work for a short moment \\ and reset again!\end{tabular}                            & \begin{tabular}[c]{@{}l@{}}`benefici', `distanc', `work', `short', \\ `moment', `reset'\end{tabular}                     \\ \hline
\end{tabular}
\end{table}

\section{Topic Modelling}\label{sec:topic_modelling}

There are various methods by which texts can be distributed into topics, with some of the most traditional methods being probabilistic Latent Semantic Analysis (pLSA) \cite{hofmann1999probabilistic}, Latent Semantic Analysis (LSA) \cite{dumais2004latent}, and LDA \cite{blei2003latent}. More recently, however, with the recent advancement in the NLP field, newly developed algorithms, such as BERTopic \cite{grootendorst2022bertopic} and Top2Vec \cite{angelov2020top2vec}, are continuing to attract attention.

Such methods vary in complexity, the representation of text segments taken as the model's input (e.g. bag-of-words, word embeddings), their computational speed, and their performance. For example, the LDA algorithm proposes a fixed number of topics in a collection of documents and assumes that each document reflects a combination of those topics. When a collection of documents are analysed under these assumptions, probabilistic inference algorithms reveal an embedded thematic structure, allowing for large collections of documents to be quickly summarised, explored, and searched \cite{blei2010probabilistic}. In general, the LDA algorithm calculates the probability that a word within a document will be included in each topic. A topic may be described by extracting words with the highest probabilities which correspond to such topic. That is, LDA analysis finds the latent topic corresponding to the words in any given document \cite{kang2019analysis}. A document is determined to address a topic by calculating a probability distribution over a range of topics for each document \cite{curiskis2020evaluation} and selects the topic with the highest probability as the main topic description. 

However, BERTopic and Top2Vec use word embeddings. That is, the vectorisation of text data makes it possible to locate semantically similar words, sentences, or documents within spatial proximity. As word vectors that emerge closest to the document vectors are considered as being the best description of topic of the document, the number of documents that can be grouped together represents the number of topics \cite{hendry2021topic}. Whereas BERTopic uses Hierarchical Dirichlet Process (HDP) to cluster vectors into topics, Top2Vec employs a combination of document clustering and word embedding techniques. It uses the Doc2Vec algorithm to generate document embeddings \cite{egger2022topic}. Unlike traditional topic modelling methods, Top2Vec does not require specifying the number of topics in advance. Instead, it identifies topic clusters based on the density of document embeddings and extracts representative keywords and documents for each topic.

Before applying the proposed approach to state-of-the-art models, initial experimentation was conducted using the traditional topic modelling method, LDA. By validating the method with LDA, which is a well-established and widely used technique, it is possible to verify that the approach functions as intended and produced satisfactory results. 

For each pre-processed dataset described in Section \ref{sec:datasets}, the topic modelling approach was applied using the `Latent Dirichlet Allocation' \cite{sklearn-lda} class available as part of the Scikit-learn package. A key hyperparameters of the LDA algorithm is the number of topics \cite{spasic2020patient}. To calculate the optimal number of topics across a collection of documents, a good indication is the number of topics with which the model best predicts the data. For topic models such as LDA, a common indicator to measure the optimal number of topics is perplexity, a measure of how well a probabilistic model predicts a sample \cite{blei2003latent}. Yet, recent studies have shown that predictive likelihood (in this case, perplexity) and human judgement are often not correlated \cite{chang2009reading}. In this case, and to retrieve a comparable number of topics, for each dataset, text responses were distributed as one of ten topics. To maintain the comparability across each dataset, the default values of the remaining hyperparameters of the LDA algorithm were set \cite{sklearn-lda}. Once topics were identified as a probability distribution of words, each text sample is denoted as a probability distribution over topics. The topic with highest probability was therefore applied to each text segment. In addition, for each topic, the top ten tokens with the highest probabilities were selected as the description of the topic. 

Dataset ID = 5 is used as an ongoing example to illustrate the results of the traditional and extended topic modelling outputs discussed in this paper. However, the results are applicable across all datasets. Table \ref{tab:top_10_words_per_topic} reports the token list output produced by the LDA approach. Evidently, based on the content of the textual responses, several words are extracted from topics that may infer its description. For example, \textit{blood donation service} may be inferred as the description for topic 1, \textit{staff mental health} may infer topic 4, and \textit{email signatures} may infer topic 8. Other topic descriptions highlight the possible challenges surrounding the inferences of topic descriptions to human readers as they include inconsistent tokens that may be ambiguous without context. Such examples include topic 3, which includes the words \textit{piano, kitchen, bond,} etc.

\begin{table}[ht]
\caption{Token outputs across topics for dataset ID = 5}
\label{tab:top_10_words_per_topic}
\centering
\begin{tabular}{|c|l|}
\hline
\textbf{Topic} & \multicolumn{1}{c|}{\textbf{LDA Output}}                                                                                                     \\ \hline
1              & \begin{tabular}[c]{@{}l@{}}`offic', `blood', `donat', `welsh', `servic', `tree',  `clinic', `onenot', `help', \\ `milk'\end{tabular}         \\ \hline
2              & \begin{tabular}[c]{@{}l@{}}`address',  `request', `item', `power', `group', `receiv', `start', `environ', \\ `real',  `differ'\end{tabular}  \\ \hline
3              & \begin{tabular}[c]{@{}l@{}}`provis',  `piano', `kitchen', `bond', `easi', `stress', `100k', `music', \\ `boost',  `consid'\end{tabular}      \\ \hline
4              & \begin{tabular}[c]{@{}l@{}}`train',  `aid', `health', `mental', `team', `staff', `provid', `quarter', `roll',  \\ `recycl'\end{tabular}      \\ \hline
5              & \begin{tabular}[c]{@{}l@{}}`calcul', `app', `mileag', `adjust', `height', `repay', `loan', `trip', `catcher',  \\ `work'\end{tabular}        \\ \hline
6              & \begin{tabular}[c]{@{}l@{}}`learn', `yoga', `repair', `lesson', `hand', `sanitis', `review', `theme', \\ `improv',  `cafe'\end{tabular}      \\ \hline
7              & \begin{tabular}[c]{@{}l@{}}`starter',  `leav', `annual', `advanc', `warn', `board', `notic', `digit', \\ `encourag',  `comm'\end{tabular}    \\ \hline
8              & \begin{tabular}[c]{@{}l@{}}`signatur', `email', `electron', `railcard', `dbw', `allow', `benefit', \\ `flexi', `lgbt',  `30'\end{tabular}    \\ \hline
9              & \begin{tabular}[c]{@{}l@{}}`mail',  `list', `websit', `vacanc', `distribut', `christma', `data', \\ `breach',  `alli', `budget'\end{tabular} \\ \hline
10             & \begin{tabular}[c]{@{}l@{}}`post',  `gener', `polici', `john', `role', `refer', `friendli', `social', \\ `messag',  `display'\end{tabular}   \\ \hline
\end{tabular}
\end{table}

\section{Extending Topic Modelling Outputs} \label{sec:keyword_extraction}

To reiterate, this paper proposes a new method towards extending the outputs of traditional topic modelling methods. The initial experiments presented herein are tested using the LDA method described in Section \ref{sec:topic_modelling}. The proposed approach includes the following steps:

\begin{itemize}

    \item Mapping the original text responses in each dataset (Table \ref{tab:preprocessing_text}) to the most dominant topic expressed in such texts by selecting the topic with the highest relevance score produced by the topic model, in this case, LDA. 
    
    \item For each topic, apply a keyword extraction approach to the original text responses.
    
    \item For each topic, map the aforementioned extracted keywords to LDA's output (Section \ref{sec:topic_modelling}) by selecting the keywords with the highest number of intersecting tokens with those produced by the LDA model. Subsequently, of those further refined keywords phrases, the top-scoring ones were assigned as the topic's main description.     
    
\end{itemize}

Keywords, which are often defined as a sequence of one or more words, provides a short description of the content of a text document \cite{rose2010automatic}. Such keywords may be useful entries for building an automatic indexing system for a document collection or can be used to classify text. In the context herein, keywords may serve as a concise label for a given topic. 

A popular keyword extraction algorithm is Rapid Automatic Keyword Extraction (RAKE) \cite{rose2010automatic}, an unsupervised, domain and language-independent tool for extracting keywords from documents. RAKE identifies stop words and phrase delimiters to split the document into candidate keywords, which are sequences of content words as they occur in the text \cite{ramanujam2016automatic}. Firstly, RAKE tokenises the document text by specific word delimiters. Using phrase delimiters and the positions of stop words, tokenised texts are then split into sequences of continuous words. Words within a sequence are assigned the same position in the text and together are considered a candidate keyword \cite{davaretext}. Once each candidate keyword is identified, a score is calculated for each and is defined as the sum of its member word scores. Such scores are calculated using one of three methods: the degree of a word in the matrix (i.e. the sum of the number of co-occurrences the word has with any other content word in the text), the word frequency (i.e. the number of times the word appears in the text), or as the degree of the word divided by its frequency. To find specific keywords, RAKE searches for pairs of keywords that adjoin one another at least twice in the same document and in the same order. A new candidate keyword is then created using a combination of the extracted keywords. The score for the new keyword is the sum of its member keyword scores. As RAKE splits candidate keywords by stop words, interior stop words are not present in the extracted keywords. 

After applying topic modelling (Section \ref{sec:topic_modelling}), the original text responses in each dataset (see Table \ref{tab:preprocessing_text}) were mapped to the most dominant topic expressed in such texts by choosing the topic with the highest relevance score. For example, for dataset = 5, topic 8 was assigned to the text response \textit{``Implementation of electronic signatures on contracts and offer letters. This will support with process improvement and digitisation''} as a consequence of achieving the highest relevant score of 0.94, whereas topic 4 was assigned to \textit{``Mental health is a very important aspect of staff wellbeing, and it affects a lot of people''} as a consequence of achieving the highest relevant score of 0.94. For each topic, RAKE was applied to the original text responses and the keywords, as well as their scores, were extracted.

The keywords extracted by RAKE were pre-processed following the techniques discussed in Section \ref{sec:datasets} and mapped to those extracted as part of the LDA's output. To facilitate this mapping, after applying direct string matching, the keywords with the highest number of intersecting tokens with those produced by the LDA model were extracted. Subsequently, of those further refined keywords phrases, the top-scoring ones were assigned as the topic's main description. Topics with more than one top-scoring keyword were concatenated and separated by the '/' delimiter. 

For example, as a consequence of including the highest number of intersecting tokens, i.e. \textit{aid, train, mental} and \textit{health} (see Table \ref{tab:top_10_words_per_topic}), as well as achieving the highest score produced by the RAKE algorithm (see Table \ref{tab:top_10_keywords}), topic 4 in dataset ID = 5 was assigned \textit{Aid Training Courses/Mental Health Issues} as the topic's descriptor. The remaining keywords may be useful in describing other topics discussed in the text responses (e.g. \textit{Mental Health Illness, Aid Training} and \textit{Training Providers}).

\begin{table}[ht]
\caption{Top scoring keyword phrases for topic 4, dataset ID = 5}
\label{tab:top_10_keywords}
\centering
\begin{tabular}{|l|c|}
\hline
\multicolumn{1}{|c|}{\textbf{RAKE Keyword Phrases}}                      & \textbf{Score} \\ \hline
Aid Training Courses & 9              \\ \hline
Mental Health Issues & 9              \\ \hline
Mental Health Illness                       & 8.33     \\ \hline
Mental Health                               & 5.33     \\ \hline
Aid Training                                & 4              \\ \hline
Training Providers                          & 4              \\ \hline
\end{tabular}
\end{table}

\begin{figure}
\centering
\includegraphics[width=1\textwidth]{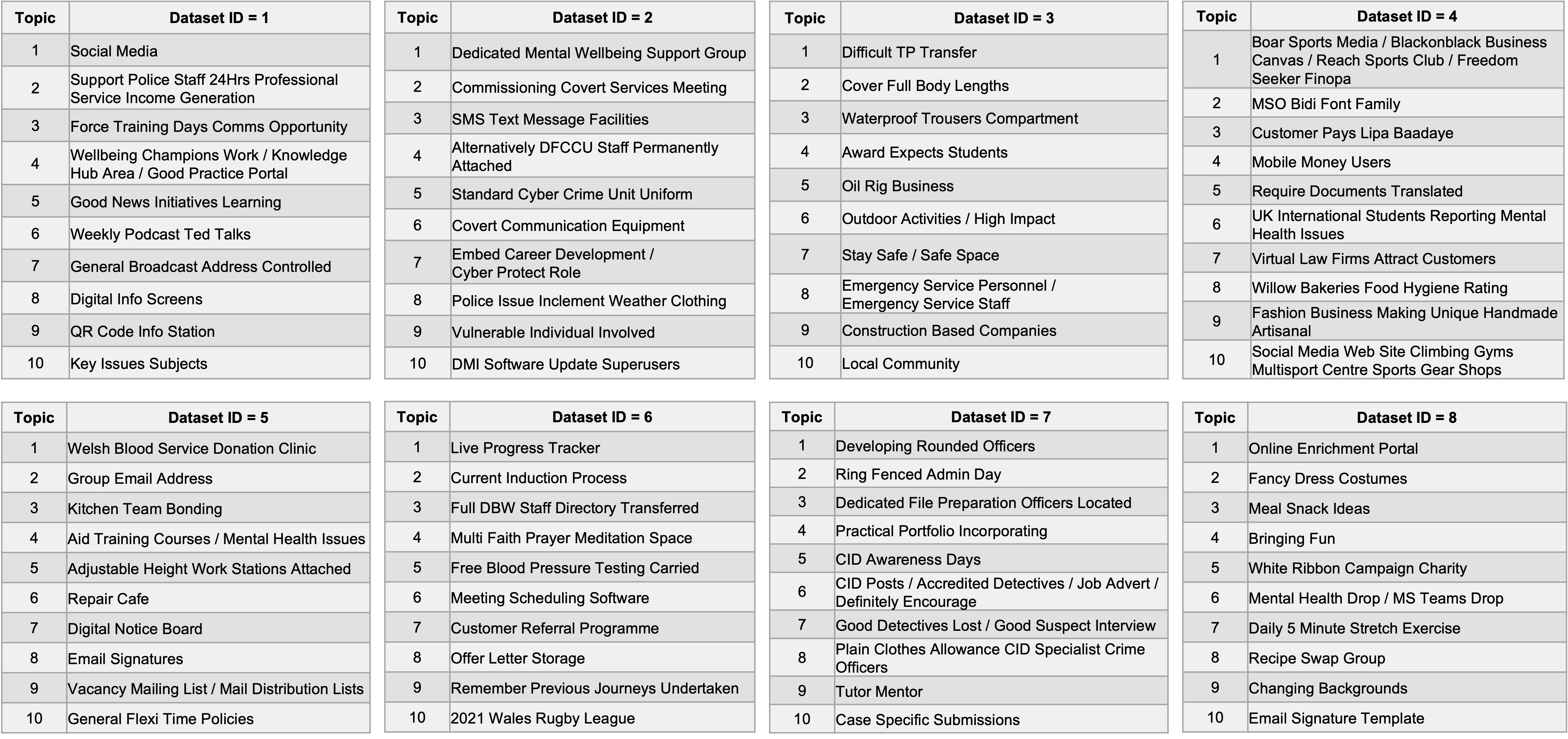}
\caption{Extended topic outputs for all datasets}
\label{tab:topic_output_vs_keyword_output}
\end{figure}

Figure \ref{tab:topic_output_vs_keyword_output} reports the extended outputs across each of the ten topic extracted for all datasets. In very few examples (e.g. \textit{Free Blood Pressure Testing Carries} (Dataset ID = 6) and \textit{Dedicated File Preparation Officers Located} (Dataset ID = 7)), extended outputs may lack minor subject-verb agreement and coherence. However, in comparison to the token list output produced by the LDA topic modelling approach (Table \ref{tab:top_10_words_per_topic}), the resulting extended method presents more cohesive and contextualised topic descriptions. 

\section{Evaluating Interpretability from a Human Perspective}\label{sec:evaluation}

Inspired by several studies discussed in Section \ref{sec:background} (e.g. \cite{kou2015automatic}, \cite{wan2016automatic}, \cite{chang2009reading}), to gain an insight into whether the proposed method of extending topic modelling outputs increases the interpretability of topic descriptions from a human's perspective, independent annotators were asked to label both topic modelling outputs according to their:

\begin{itemize}
    \item \textbf{Quality} - how easy is it to extract meaning from the text?
    
    \item \textbf{Usefulness} - how relevant or helpful is the text in providing information about a topic?
    
    \item \textbf{Efficiency} - how efficient is the annotation task?
\end{itemize}

\begin{figure}
\includegraphics[width=1\textwidth]{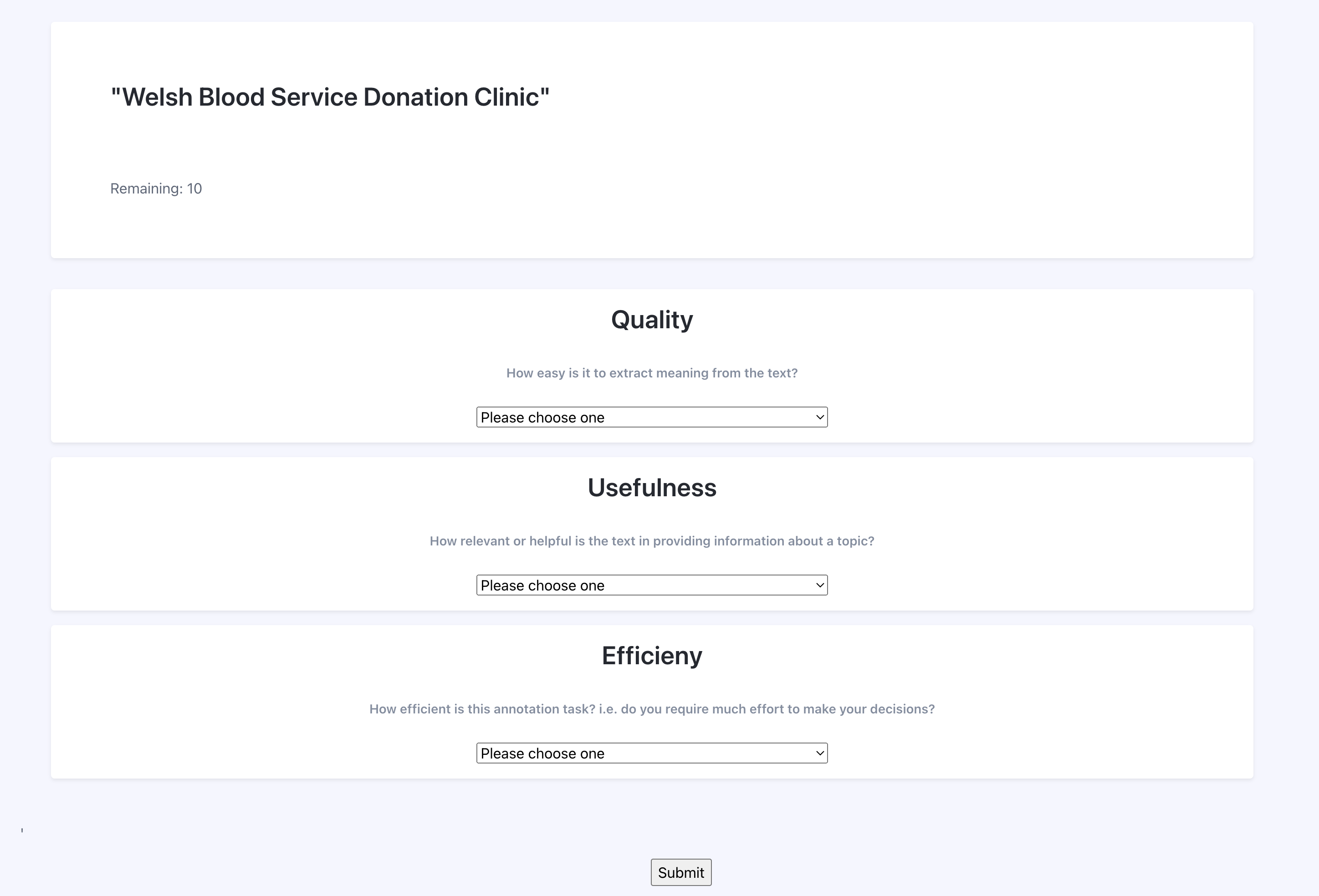}
\caption{Bespoke annotation platform}
\label{annotation_platform}
\end{figure}

To facilitate the annotation task, a bespoke web-based annotation platform was implemented. This reduced any installation overhead and widened the reach of annotators as it was accessible using a web browser regardless of the device type (i.e. smartphone, laptops, PC, etc.). Annotators were presented with instructions explaining the task's requirements and then with the platform's interface (see Figure \ref{annotation_platform}. The first pane contained a randomly selected topic modelling output to be annotated, as well as the remaining number of annotations left to complete. For the LDA outputs, to increase interpretability, annotators were presented with non-stemmed tokens. The subsequent panes contained the annotation choices for the aforementioned metrics. For each metric, annotators were required to label each output on a 5-point Likert scale, where a score of 0 signified poor quality, no usefulness, and not an efficient task, and a score of 4 signified high quality descriptions containing clear and effective text that is easy to extract meaning from, are useful in describing a topic, and they are performing the annotation task in the best possible manner with the least amount of effort.   

The crowdsourcing of labelling natural language often use a limited number of annotators with the expectation that they are perceived to be experts \cite{tarasov2010using}. However, the task of annotating text is considered as being highly subjective and varies with the annotator's age, gender, experience, cultural location, and individual psychological differences \cite{passonneau2008relation}. For instance, Snow et al. \cite{snow2008cheap} investigate collecting annotations from a wide audience of non-expert annotators over the Web. They show high agreement between the 10 annotations provided by non-experts and those provided by experts. 

In this case, to crowdsource annotations in this study and to reach a diverse range of annotators, Twitter was leveraged as a distribution channel. The study was able to attract a broad audience interested in contributing to the annotation process. Annotators were not selected selectively; instead, participation was open to anyone who expressed interest in contributing their insights. To maintain anonymity and avoid duplication of annotations, annotators were distinguished solely by their IP addresses. No personal information was collected from annotators, as explained in the privacy policy, ensuring the confidentiality of their identities. All annotation results were securely stored in a relational database.

A total of 1,600 annotations were collected for all 80 outputs described in Section \ref{sec:topic_modelling} and 80 extended outputs described in Section \ref{sec:keyword_extraction} with 10 annotations per output. A total of 63 annotators participated in the study.

\begin{figure}
\includegraphics[width=1\textwidth]{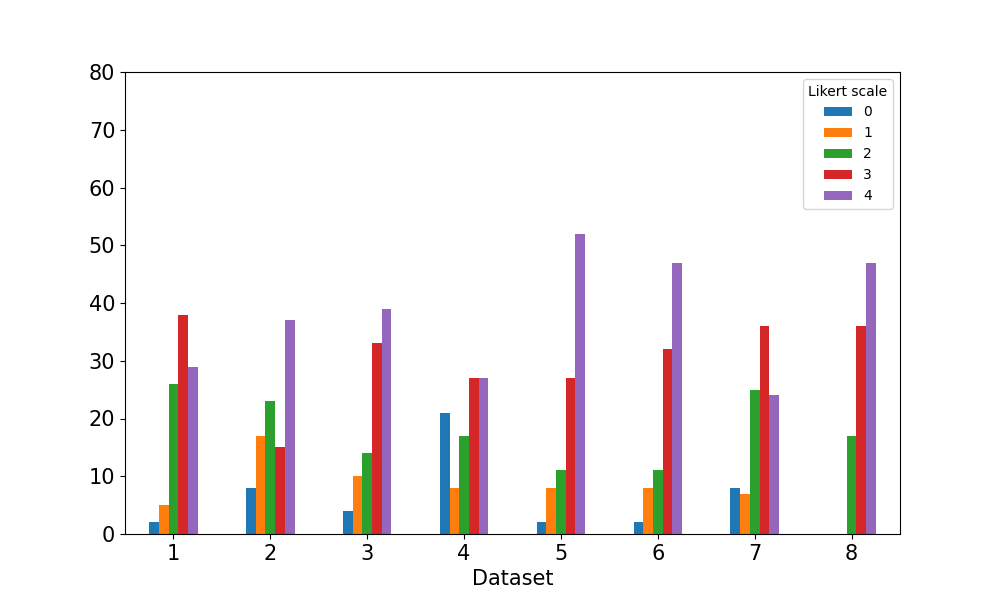}
\caption{Distribution of annotations across the quality of extended outputs}
\label{distribution_of_quality_annotations_extended}
\end{figure}

\begin{figure}
\includegraphics[width=1\textwidth]{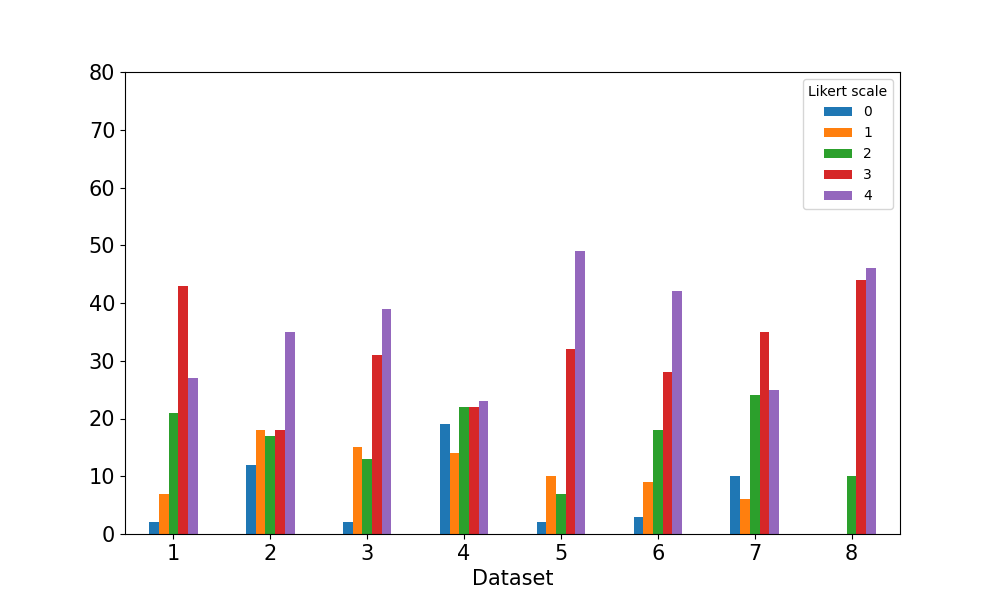}
\caption{Distribution of annotations across the usefulness of extended outputs}
\label{distribution_of_usefulness_annotations_extended}
\end{figure}

\begin{figure}
\includegraphics[width=1\textwidth]{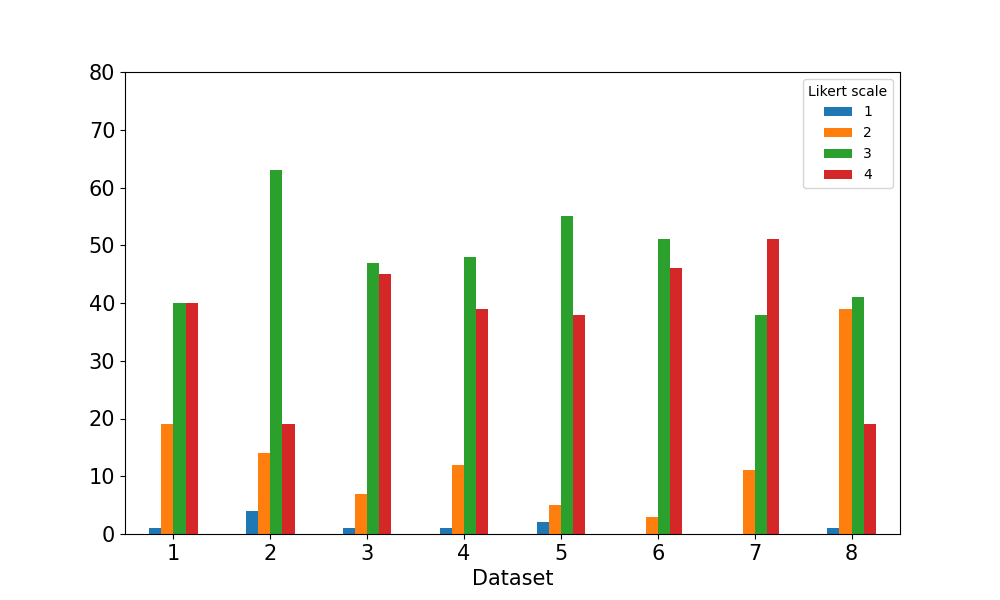}
\caption{Distribution of annotations across the efficiency of extended outputs}
\label{distribution_of_efficiency_annotations_extended}
\end{figure}

\begin{figure}
\includegraphics[width=1\textwidth]
{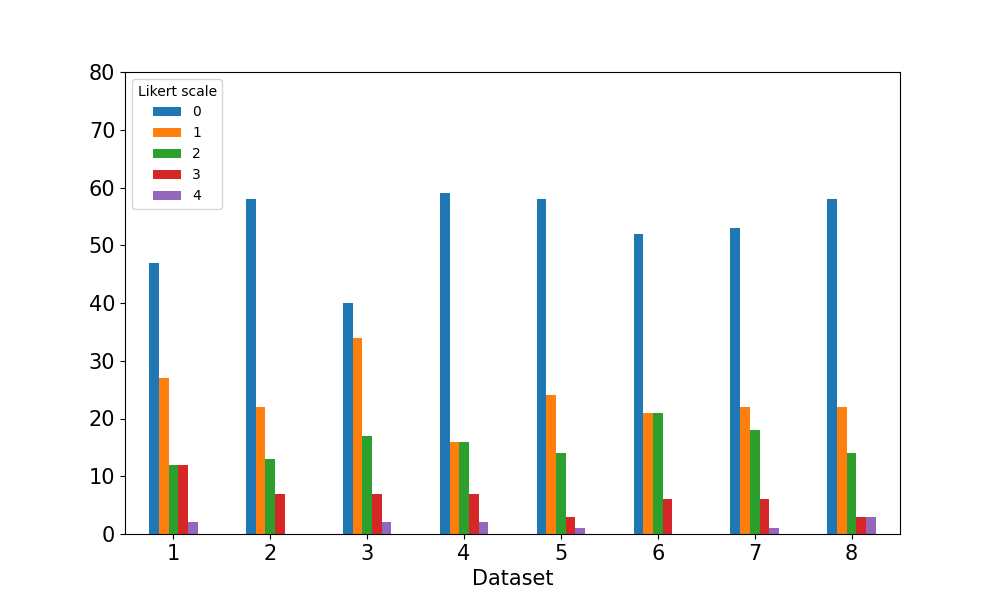}
\caption{Distribution of annotations across the quality of LDA outputs}
\label{distribution_of_quality_annotations_lda}
\end{figure}

\begin{figure}
\includegraphics[width=1\textwidth]{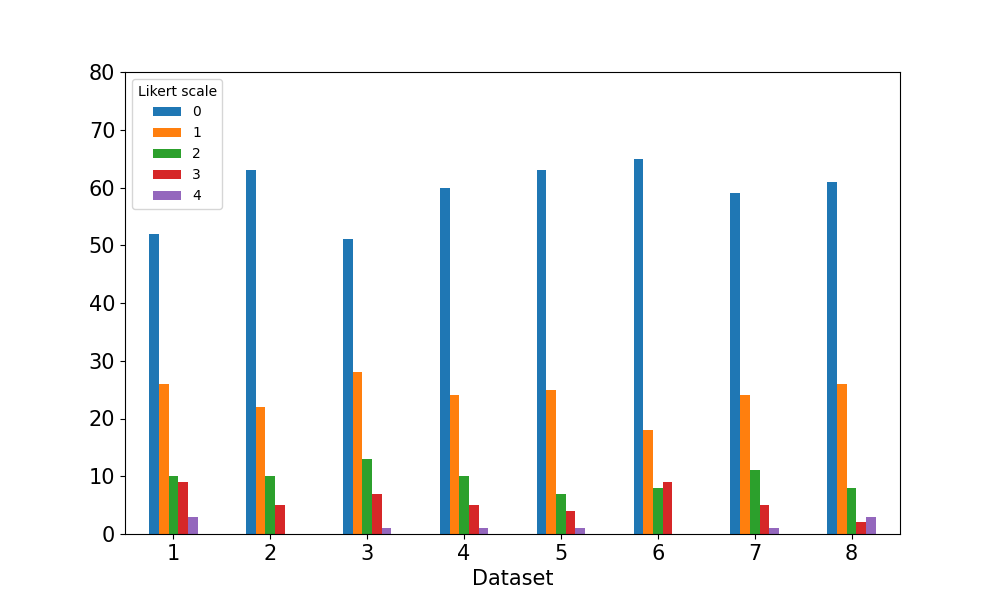}
\caption{Distribution of annotations across the usefulness of LDA outputs}
\label{distribution_of_usefulness_annotations_lda}
\end{figure}

\begin{figure}
\includegraphics[width=1\textwidth]{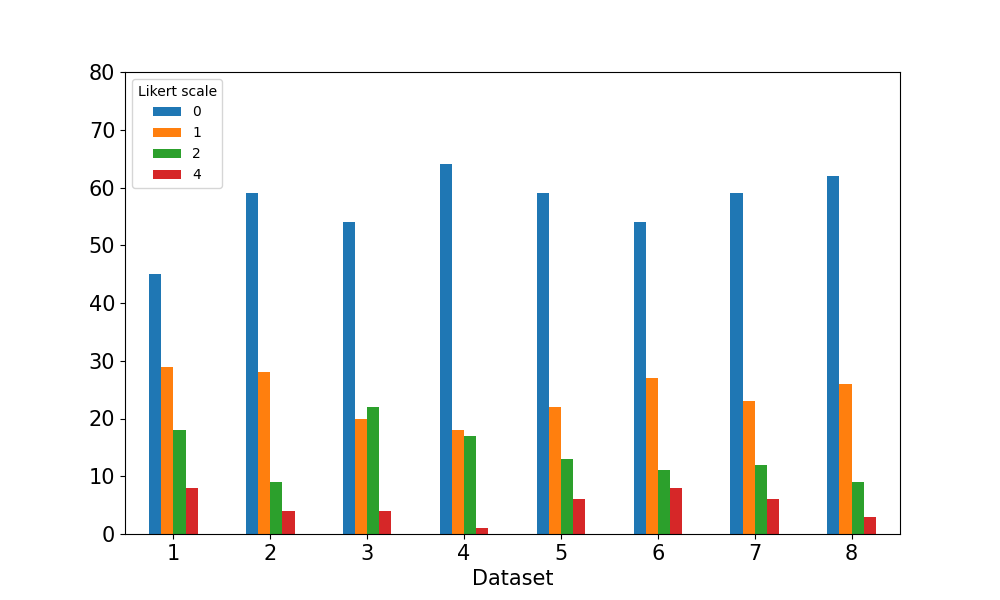}
\caption{Distribution of annotations across the efficiency of LDA outputs}
\label{distribution_of_efficiency_annotations_lda}
\end{figure}

The distributions of annotations across quality, usefulness, and efficiency scales for extended topic modelling outputs across each dataset described in Table \ref{tab:distribution} are shown in Figures \ref{distribution_of_quality_annotations_extended}-\ref{distribution_of_efficiency_annotations_extended} respectively, whilst Figures \ref{distribution_of_quality_annotations_lda}-\ref{distribution_of_efficiency_annotations_lda} report the distributions of annotations across LDA outputs. For both quality and usefulness, the extended outputs incurred by far a highest usage of 4 on the Likert scale in comparison to the LDA outputs (quality = 302 out of 800 (37.8\%), usefulness = 286 out of 800 (35.8\%)). An interesting observation is that dataset ID = 8 did not receive any annotation scores of 0 or 1 for both metrics. Conversely, the LDA outputs incurred by far the highest usage of 0 (quality = 425 out of 800 (53.1\%), usefulness = 474 out of 800 (59.3\%)). Further analysis demonstrates that datasets ID = 2 and 6 did not receive any annotations of scores of 4 for both metrics. 

In terms of task efficiency, similar results can be observed. For the extended outputs, a large distribution of 3 and 4 on the Likert scale were used by annotators, denoting that the task of interpreting the extended topic descriptions did not require a significant amount of effort to complete. On the other hand, the LDA outputs report to have a high distribution of 0 on the Likert scale, indicating that annotators required much more effort in defining whether the quality and usefulness of the topic descriptions.

\begin{figure}
\includegraphics[width=1\textwidth]{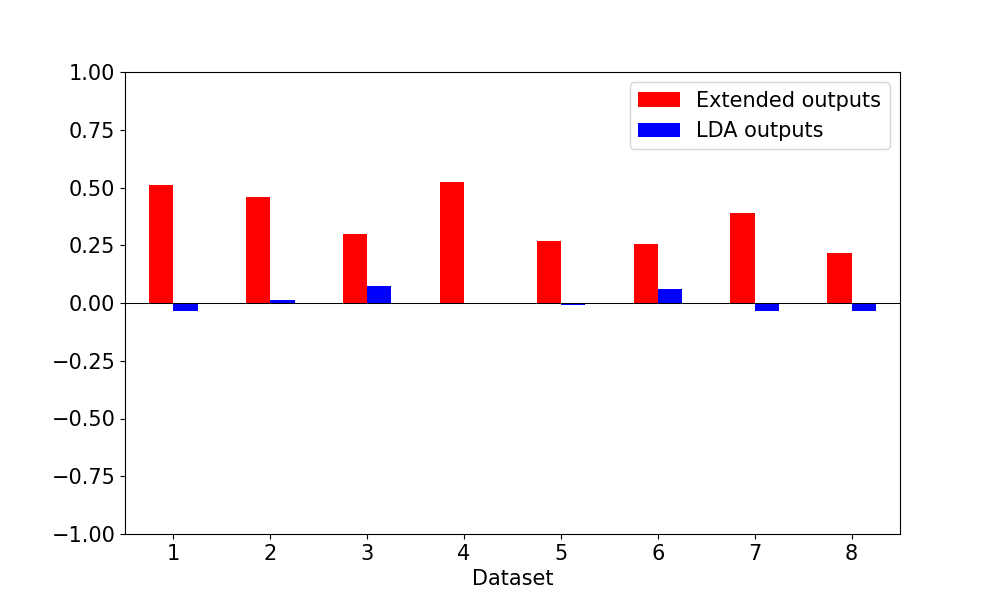}
\caption{Inter-annotator agreement across quality}
\label{quality_agreement_results}
\end{figure}

\begin{figure}
\includegraphics[width=1\textwidth]{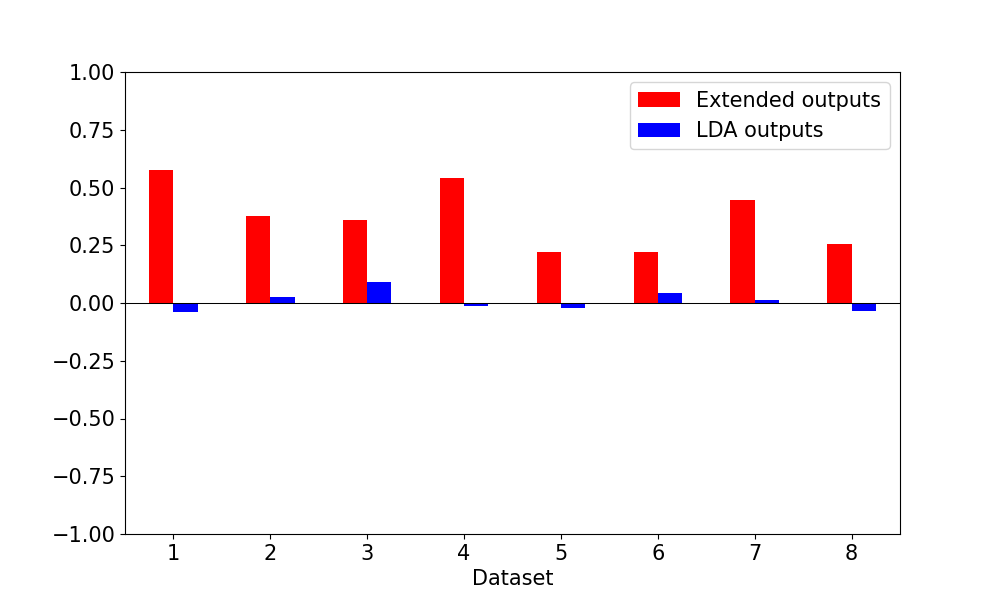}
\caption{Inter-annotator agreement across usefulness}
\label{usefulness_agreement_results}
\end{figure}

\begin{figure}
\includegraphics[width=1\textwidth]{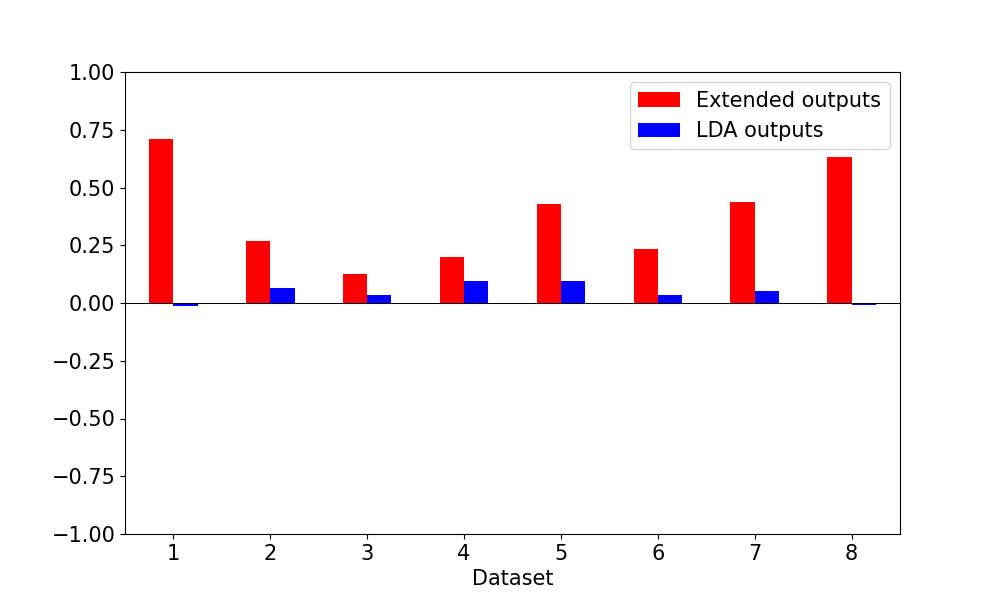}
\caption{Inter-annotator agreement across efficiency}
\label{efficiency_agreement_results}
\end{figure}

We hypothesise that when an output is unambiguous and helpful in providing information about a topic, then the likelihood of independent annotators selecting higher scores for the metrics increases. This leads to a higher inter-annotator agreement which indicates that a topic output is more interpretable. Krippendorff's alpha coefficient \cite{krippendorff2018content} was used to measure the inter-annotator agreement. As a generalisation of known reliability indices, it was used as it applies to: (1) any number of annotators, not just two, (2) any number of categories, and (3) corrects for chance expected agreement. Krippendorff's alpha coefficient of 1 represents full agreement, 0 represents no agreement beyond chance, and -1 represents disagreement.

Figures \ref{quality_agreement_results}-\ref{efficiency_agreement_results} report the inter-annotator agreement for the quality and usefulness of both the topic modelling outputs across each dataset, as well as the efficiency of the annotation task. For quality, the extended outputs reported the highest agreement of $\alpha$ = 0.522 (dataset ID = 4) and the lowest agreement of $\alpha$ = 0.216 (dataset ID = 8). Likewise, for usefulness, the extended outputs reported the highest agreement of $\alpha$ = 0.575 (dataset ID = 1) and the lowest agreement of $\alpha$ = 0.220 (dataset ID = 5 and 6). In terms of the LDA outputs, for quality, the highest agreement reported was $\alpha$ = 0.075 (dataset ID = 3) and the lowest agreement was $\alpha$ = -0.036 (dataset ID = 7). Likewise, for usefulness, the LDA outputs reported the highest agreement of $\alpha$ = 0.090 (dataset ID = 3) and the lowest agreements of $\alpha$ = -0.040 (dataset ID = 1). 

For extended outputs, the relatively high agreement for both quality and usefulness may be explained by the fact that they received a higher distribution of higher scoring annotations. For instance, referring back to the ongoing example of dataset ID = 5 (see Table \ref{tab:topic_output_vs_keyword_output}), the extended output \textit{Group Email Address} received an unanimous agreement of a score of 4 across both quality and usefulness. Whereas due to the ambiguity of the topic description given by the LDA output (``\textit{address, request, item, power, group, receive, start, environment, real, differ''}), it received the following distribution of annotations: quality (0 = 8 annotations, 1 = 1 annotation, 2 = 1 annotation) and usefulness (0 = 8 annotations, 1 = 2 annotations). 

In terms of task efficiency, similar agreements can be found. It was reported, for extended outputs, the highest agreement of $\alpha$ = 0.709 (dataset ID = 1) and the lowest agreement of $\alpha$ = 0.126 (dataset ID = 3). For the LDA outputs, the highest agreement of $\alpha$ = 0.094 (dataset ID = 4) and the lowest agreement of $\alpha$ = -0.012 (dataset ID = 1) were reported. Overall, the results demonstrate that the extended outputs are reliably more interpretable to human readers compared to LDA outputs, and the task of annotating their quality and usefulness is particularly less challenging for annotators when observing the extended outputs in comparison to LDA outputs.

\section{Generalisation of the Proposed Method using other Topic Modelling Approaches}\label{sec:generalisation}

The initial experiments conducted with LDA provided a solid foundation, demonstrating the effectiveness of the approach in generating intepretable topic descriptors. Building upon this, to demonstrate the generalisability and robustness of the proposed approach, the methodology is extended to include BERTopic and Top2Vec, two advanced state-of-the-art topic modelling techniques.

\begin{figure}
\centering
\includegraphics[width=1\textwidth]{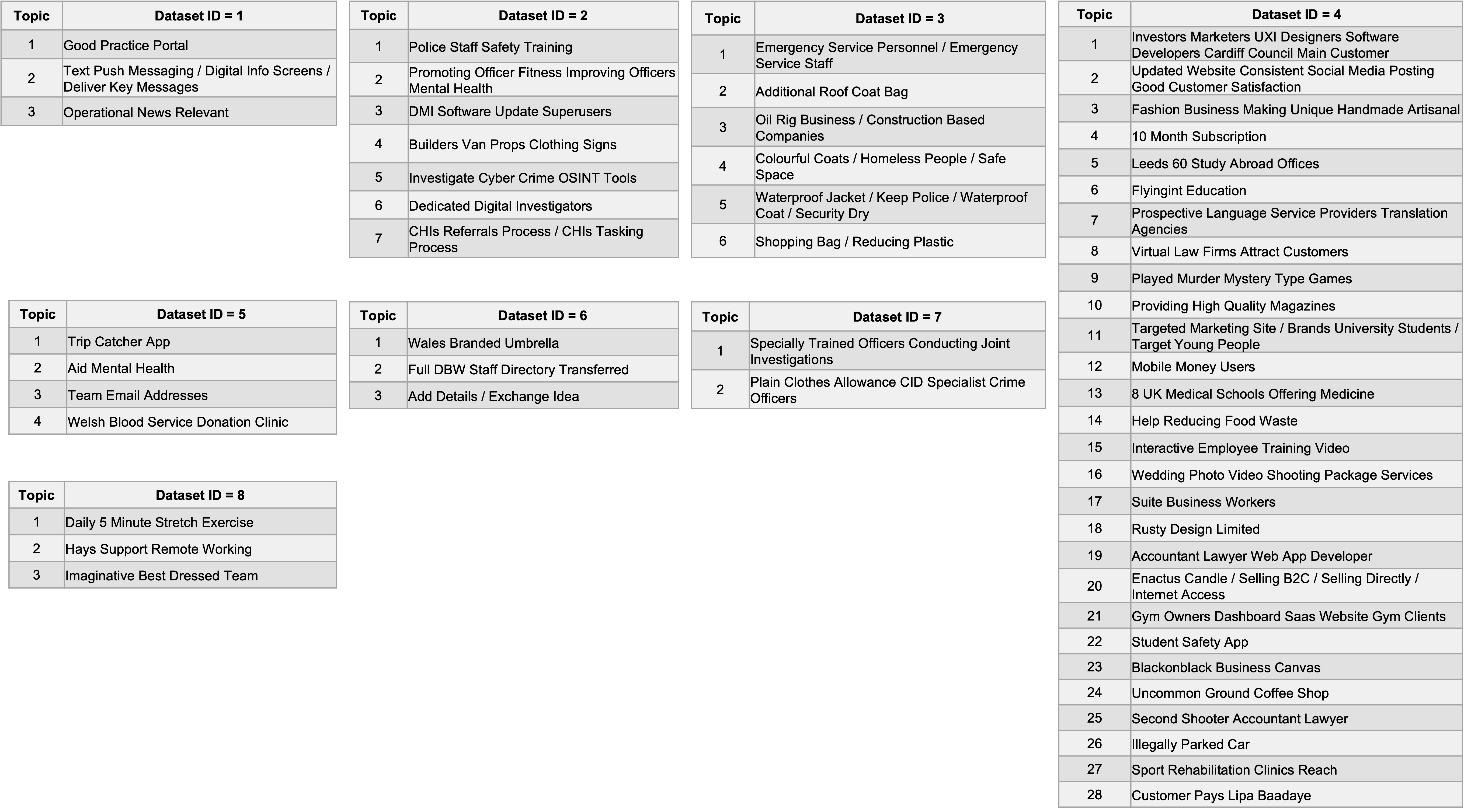}
\caption{Extended topic outputs for all datasets using BERTopic as the topic model}
\label{tab:BERTopic}
\end{figure}

\begin{figure}
\centering
\includegraphics[width=1\textwidth]{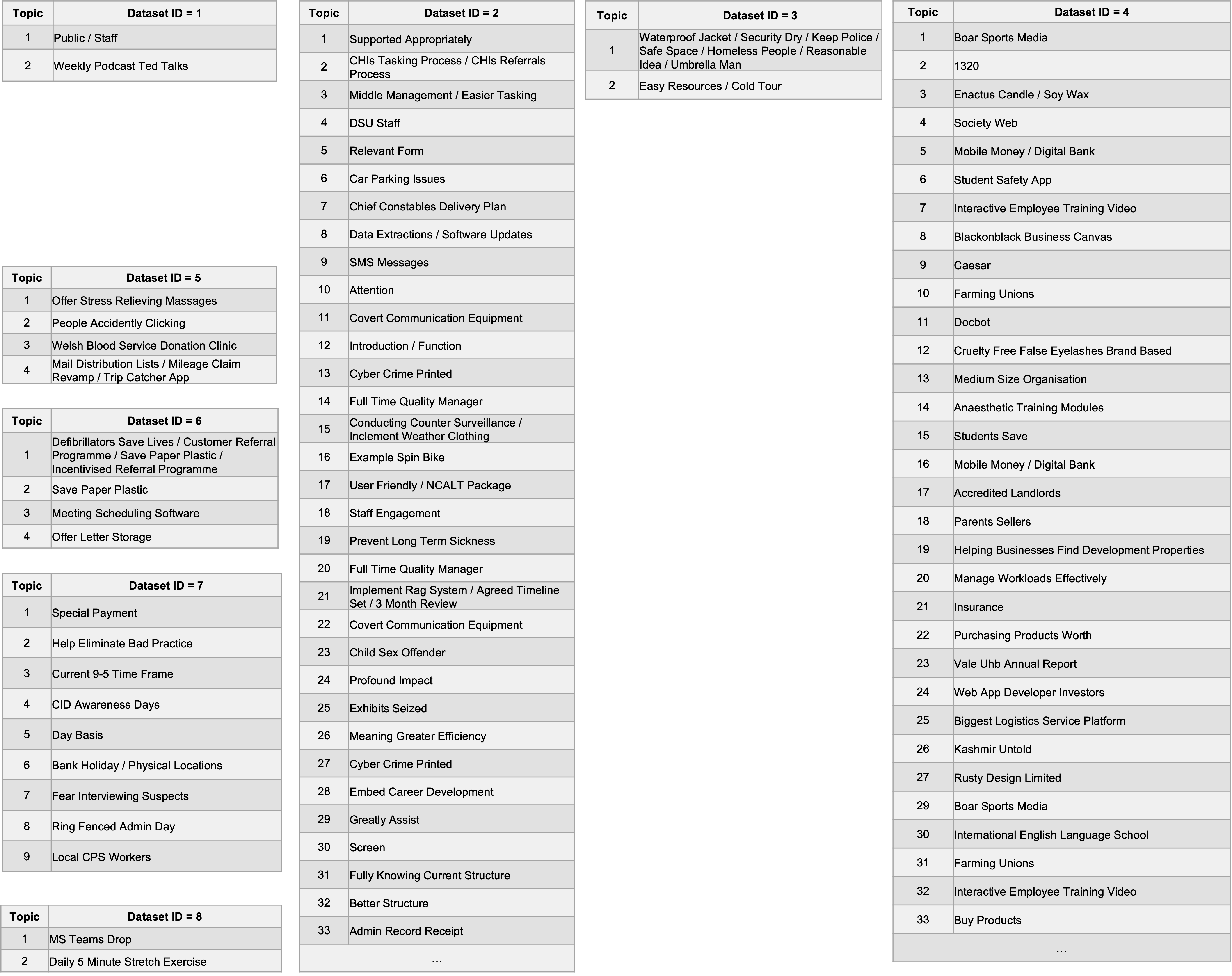}
\caption{Extended topic outputs for all datasets using Top2Vec as the topic model}
\label{tab:top2vec}
\end{figure}

Figures \ref{tab:BERTopic} and \ref{tab:top2vec} report the extended outputs across each of the topics extracted for all datasets when BERTopic and Top2Vec were used as the underlying topic modelling method, respectively. Both methods were applied without any constraint on the number of topics to be generated.

The implementation of BERTopic and Top2Vec present a broader spectrum of insights. BERTopic, with its deep learning foundation, offered a more nuanced understanding of the text data, generating topic descriptions that were not only coherent but also contextually rich. This method was particularly effective in capturing the subtleties and complexities within the datasets, which LDA might have overlooked due to its probabilistic nature. For instance, in Dataset ID = 4, BERTopic provided a multi-dimensional view of topics, reflecting a deeper layer of thematic understanding.

Top2Vec, on the other hand, presented a unique perspective by generating an extensive array of topics, exceeding 100 in Dataset ID = 4. This method's ability to produce such a wide range of topics highlights its utility in exploring large and diverse datasets, offering a comprehensive landscape of themes and ideas present in the data. However, it's important to note the presence of single-word topic descriptions in some cases. While these may seem less informative at first glance, they play a crucial role in offering a focus on specific themes, as seen with terms like \textit{Insurance} and \textit{Docbot}.

The comparative analysis of topic descriptions across LDA, BERTopic, and Top2Vec revealed interesting patterns. The consistency in topic descriptions across different methods, as shown in Dataset ID = 8 with descriptions like \textit{Daily 5 Minute Stretch Exercise}, underscores the reliability of the approach. This consistency is also an affirmation of the underlying thematic structures within the datasets. Furthermore, the alignment of topic descriptions across various models indicates a convergence of thematic interpretation, despite the inherent differences in the methodologies the topic models. This congruence suggests that this approach is capable of capturing the core essence of topics regardless of the underlying topic modelling technique used.

\section{Generalisation of the Proposed Method using New Data}\label{sec:generalisation_data}

The initial experiments conducted herein were with a private dataset provided by our industrial research collaborator. To demonstrate the generalisability of the proposed approach of extending topic model descriptors, the method is applied when using an unseen dataset.

The 20 Newsgroups dataset \cite{lang20newsgroups} is commonly used for text mining applications. It is a collection of approximately 20,000 news group documents, distributed (nearly) evenly across 20 different news groups. The documents within the 20 Newsgroups dataset exhibit a range of lengths and cover a spectrum of topics, from religion to politics to automotive and technological discussions, providing a comprehensive basis for testing the robustness of the proposed approach in a more heterogeneous textual environment, enhancing the external validity of this research. Table \ref{20newsgroups} shows how the 20 news groups are partitioned according to subject matter.

\begin{table}[ht]
\caption{20 Newsgroup dataset categories}
\label{20newsgroups}
\centering
\begin{tabular}{|l|l|l|}
\hline
\begin{tabular}[c]{@{}l@{}}comp.graphics\\ comp.os.ms-windows.misc\\ comp.sys.ibm.pc.hardware\\ comp.sys.mac.hardware\\ comp.windows.x\end{tabular} & \begin{tabular}[c]{@{}l@{}}rec.autos\\ rec.motorcycles\\ rec.sport.baseball\\ rec.sport.hockey\end{tabular} & \begin{tabular}[c]{@{}l@{}}sci.crypt\\ sci.electronics\\ sci.med\\ sci.space\end{tabular}         \\ \hline
misc.forsale                                                                                                                                        & \begin{tabular}[c]{@{}l@{}}talk.politics.misc\\ talk.politics.guns\\ talk.politics.mideast\end{tabular}     & \begin{tabular}[c]{@{}l@{}}talk.religion.misc\\ alt.atheism\\ soc.religion.christian\end{tabular} \\ \hline
\end{tabular}
\end{table}

\begin{figure}
\centering
\includegraphics[width=0.9\textwidth]{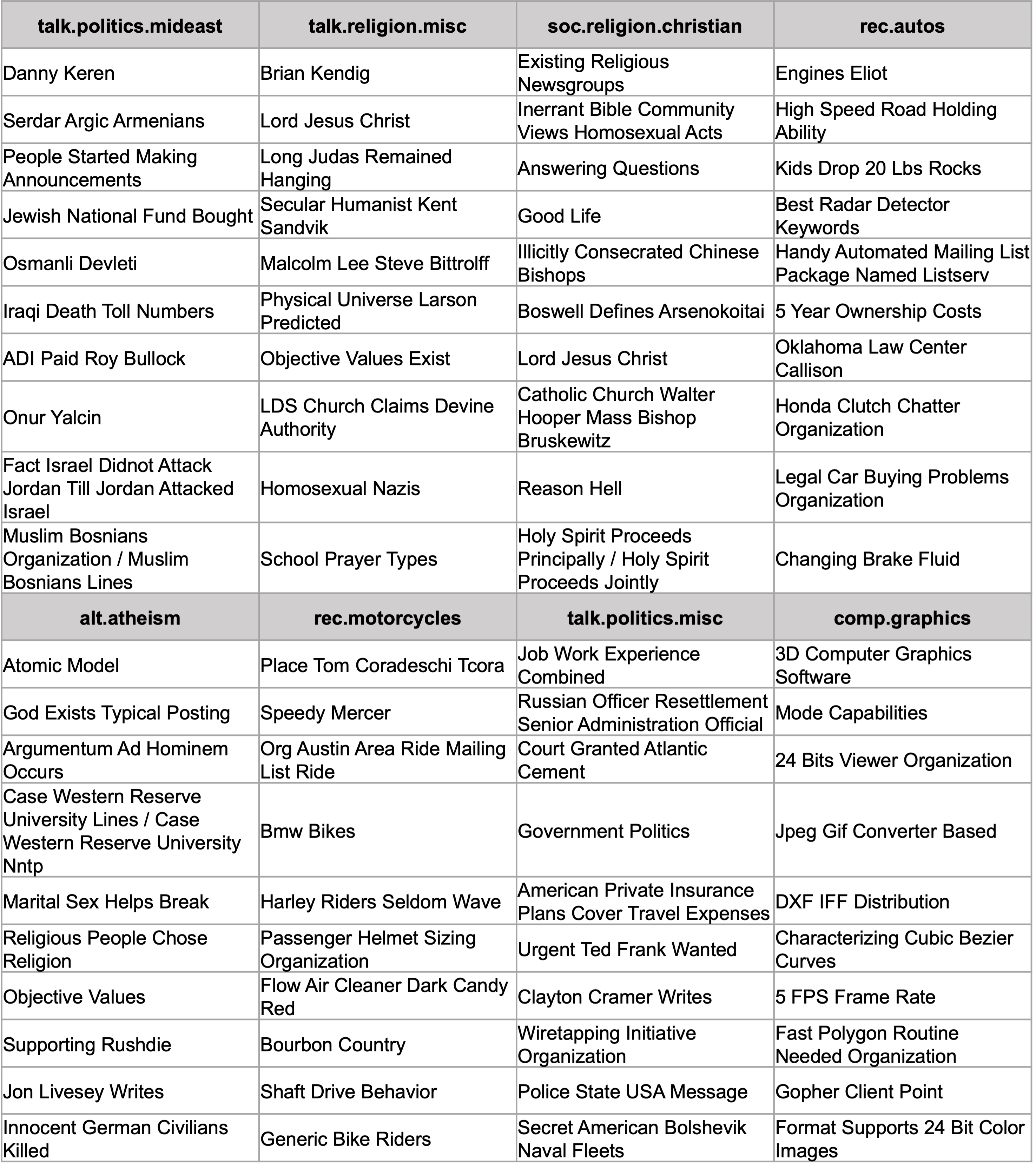}
\caption{An excerpt of the extended topic outputs for the 20 Newsgroups dataset using LDA as the topic model}
\label{tab:lda20}
\end{figure}

\begin{figure}
\centering
\includegraphics[width=0.9\textwidth]{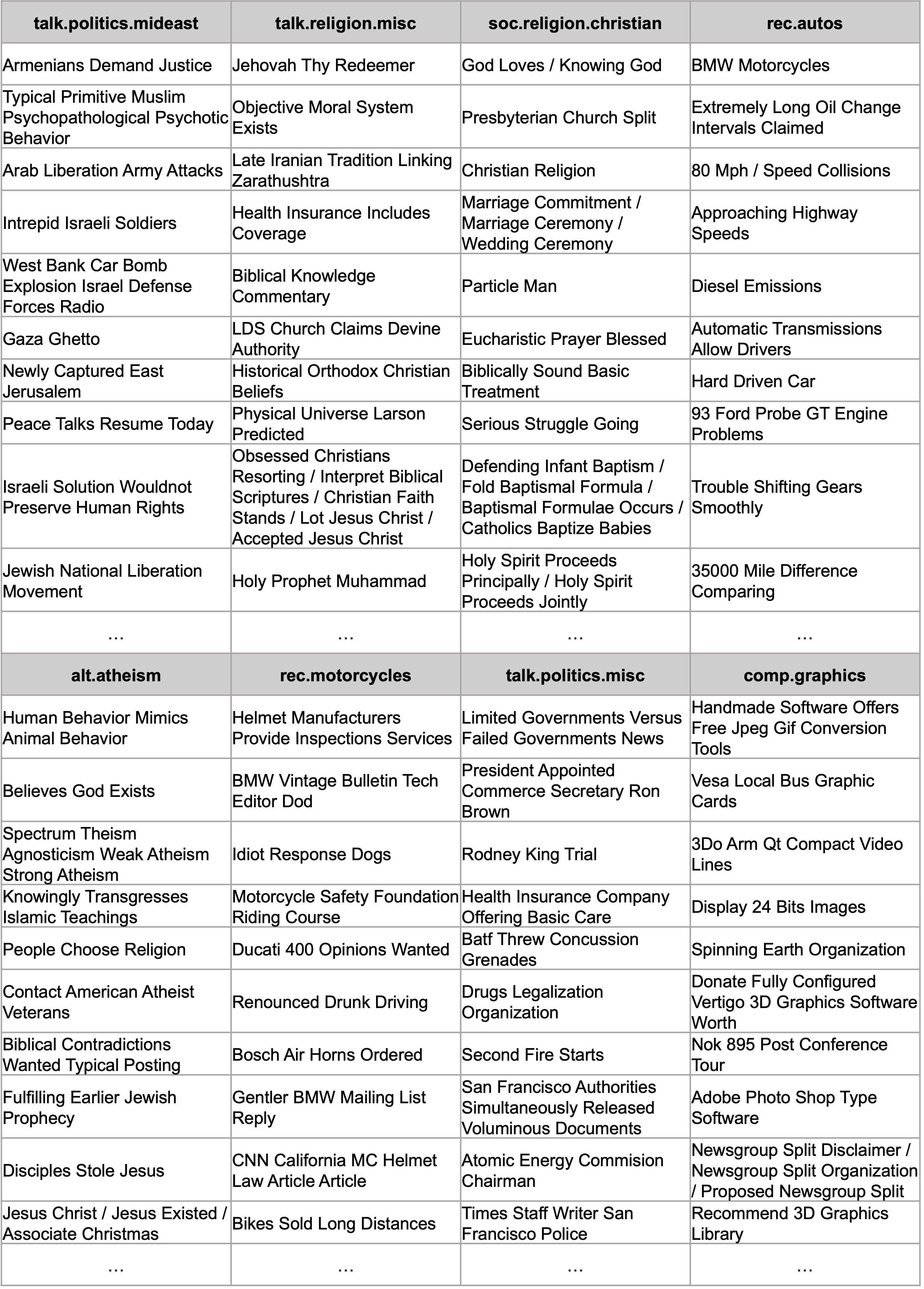}
\caption{An excerpt of the extended topic outputs for the 20 Newsgroups dataset using BERTopic as the topic model}
\label{tab:BERT20}
\end{figure}

\begin{figure}
\centering
\includegraphics[width=0.9\textwidth]{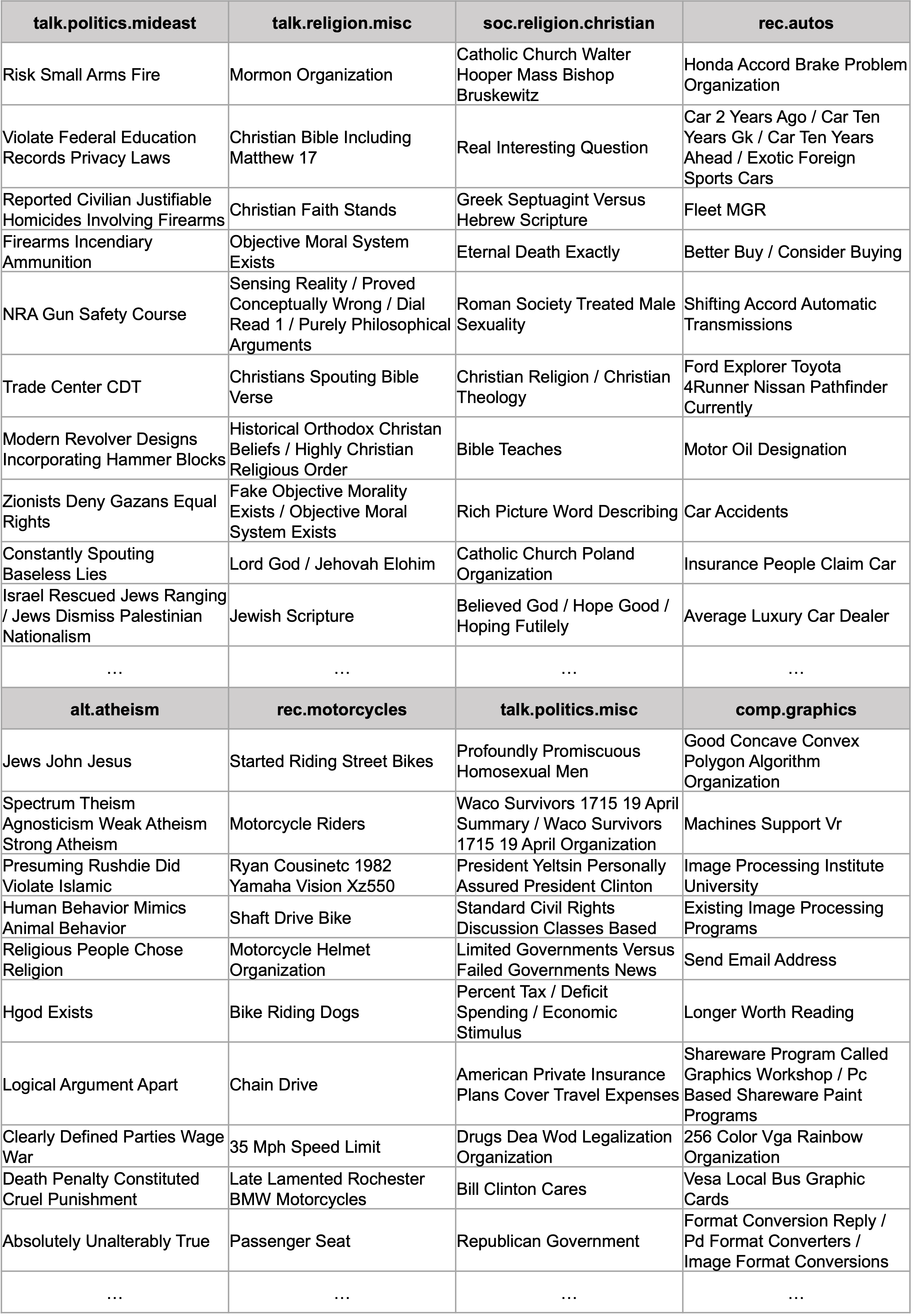}
\caption{An excerpt of the extended topic outputs for the 20 Newsgroups dataset using Top2Vec as the topic model}
\label{tab:top2vec20}
\end{figure}

Figures \ref{tab:lda20}-\ref{tab:top2vec20} report an excerpt of categories from the 20 Newsgroup dataset, where the extended outputs across each of the topics extracted when the LDA, BERTopic, and Top2Vec methods were used as the underlying topic modelling method, respectively. For LDA, the model was requested to distribute texts from each category into one of 10 topics. The BERTopic and Top2Vec were applied without any constraint on the number of topics to be generated.

Upon reapplication of the proposed method to the 20 Newsgroups dataset, the extended descriptors illustrate coherent and distinctive topics that align with known categories within the dataset. For instance, when using all three models, it is reported that topics related to 'talk.politics.mideast' are accurately associated with region-specific political discussions (e.g. \textit{Jewish National Liberation Movement, Iraqi Death Toll Numbers}), while categories like 'soc.religion.christian' and 'rec.autos' have keywords strongly related to religious discourse (e.g. \textit{Lord Jesus Christ, Jehovah Thy Redeemer}) and automotive subjects (e.g. \textit{Changing Brake Fluid, BMW Motorcycles}), respectively. These outputs also demonstrate a nuanced understanding of the dataset, capturing the finer subtleties and variations within the broader topics. For example, they distinguish between different facets of religious discussion, separating general Christian talk from more specific debates around biblical interpretation. Similarly, in the automotive category, a separation between general automotive discussions and more technical conversations about car maintenance and mechanical issues may be observed.

\section{Limitations}\label{sec:limitations}

The proposed approach utilises a safeguard against the scenario where a text document may not contain a keyword that accurately encapsulates its topic. As keyword extraction is employed on a collective set of texts associated with a topic rather than on individual documents, the likelihood of not finding a representative keyword is significantly reduced. The aggregation of texts for each topic increases the probability that relevant keywords will emerge from the collective context, rather than relying on a singular document to provide them. Moreover, the intersectionality between tokens generated by topic modelling methods and extracted keywords further refines the relevance, as only those keywords sharing common tokens with the topic modelling output are considered. This filtration ensures that the final topic descriptions are not only grounded in the statistically derived patterns of the topic model, but are also contextualised by the actual language usage within the texts. However, despite the robustness of this method and its effectiveness in providing topic descriptions for all topics, there remains a possibility that the keywords may not always perfectly match the topic modelling output. In instances where a clear and representative keyword is not identified, or if the extracted keywords do not align well with the modeled topics, such topics can be labeled as `miscellaneous'. This categorisation serves as an acknowledgment of the method's limitations and provides an opportunity for further exploration and analysis, possibly involving manual review or alternative methods, to better understand and describe these less straightforward topics.

In addition, while the proposed method of expanding topic descriptors in this paper has shown to be effective when using different underlying topic modelling approaches, as well as when it is applied on unseen data, the limitation of this study is the comparison of the proposed approach against relevant baselines discussed in Section \ref{sec:background}. As noted in Section \ref{sec:background}, other proposed methods are not generalised and are instead tailored to specific datasets and resources, supporting topic extraction from particular genres of text, such as legal documents. These methods also often aim to provide generalised terms for topics based on the specific nature and source of the texts, rather than offering a broad application across various domains. This specialisation limits the comparability of these methods with the more generalised approach proposed in this study. Thus, while a comparison with these methods would be informative, the distinct methodologies and domain-specific applications make direct comparisons complex and perhaps not entirely equitable.

\section{Conclusion}\label{sec:conclusion}

This paper presents an approach towards extending the output of traditional topic modelling methods beyond a list of isolated tokens. The proposed approach removes the dependence on external sources by using the textual data itself by extracting high-scoring keywords and mapping them to the topic modelling method's token outputs. This approach, in turn, increases the interpretability of topic descriptions for human readers by allowing more context and information to be identified, which is an important factor to consider in use cases such as decision making and information retrieval.  

To support the experiments presented herein, eight datasets containing short textual responses were used. Once each dataset was pre-processed, the LDA algorithm was applied and set to distribute each text response across one of ten topics. For each topic, the keyword extraction algorithm, RAKE, was applied to the original text responses to extract important keywords and phrases. The token outputs generated by the LDA were subsequently mapped to outputs extracted by RAKE. The keywords with the highest number of intersecting tokens with those produced by the LDA model, as well as those that achieved the highest score from RAKE, were assigned as the topic's main description. 

Under the hypothesis that the extended topic modelling outputs are more interpretable than the LDA algorithm's output, such results were expected to be associated with high inter-annotator agreement across interpretability scores. Krippendorff’s alpha coefficient was used to measure inter–annotator agreement according to which the extended outputs achieved higher agreement for quality ($\alpha$ =  0.522) and usefulness ($\alpha$ = 0.575). Whereas the LDA outputs achieved lower agreement for quality ($\alpha$ =  0.075) and usefulness ($\alpha$ = 0.090). In terms of the efficiency of the annotation task, the extended outputs achieved higher agreement ($\alpha$ =  0.709) in comparison to the LDA outputs ($\alpha$ = -0.012). Subsequently, the analysis highlights that proposed approach increases the interpretability of the topic model results from a human perspective. To demonstrate and reinforce the generalisation of the proposed method, the approach was further applied using two of the latest state-of-the-art topic modelling methods, BERTopic and Top2Vec, using a heterogeneous unseen dataset.

\section{Future Work}\label{sec:future_work}

While the current method focuses on making topic descriptors more interpretable to human readers, it is crucial to determine whether these descriptors accurately reflect the underlying content of the topics. To assess this, as part of future work, a combination of qualitative and quantitative methods can be employed. Qualitatively, user studies can be conducted where domain experts and potential users evaluate the relevance and clarity of the topic descriptors. Quantitatively, statistical methods such as coherence measures, which assess the degree of semantic similarity within the topics, can be applied to evaluate the precision of the descriptors. Furthermore, computational experiments comparing the descriptors with unaltered topic outputs could provide quantitative insights into their accuracy and utility. Conducting this further work may offer valuable insights into optimising or fine-tuning the current methodology. This comparative analysis could highlight specific areas where the current approach may be adapted or improved to better suit different text genres or data sources. Additionally, such a study might also pave the way for developing a new, more robust method for expanding topic modeling descriptors. This potentially new method could incorporate the strengths of both generalised and specialised approaches, offering a more versatile solution that can be effectively applied across all domains and text types. By exploring these possibilities, we aim not only to refine our current approach but also to contribute to the broader field of topic modeling, enhancing its applicability and effectiveness in various contexts.

\section*{Acknowledgements}

The authors are grateful to Dr Will Webberly and Dr John Barker, our industry collaborators, from SimplyDo Ideas, Cardiff, for providing initial data to facilitate the experiments herein, as well as providing informal feedback on the interpretability of the propose approach towards extending topic model descriptors.

\bibliographystyle{unsrt}  
\bibliography{references}  






\end{document}